%% file: root.tex
\documentclass[letterpaper, 10 pt, conference]{ieeeconf}  

\IEEEoverridecommandlockouts                              

\usepackage{graphicx}
\usepackage{amsmath}
\usepackage{amssymb}
\usepackage{booktabs}
\usepackage{multirow}
\usepackage{multicol}
\usepackage{comment}
\usepackage{footnote}
\usepackage{bbding}
\usepackage{pifont}
\usepackage{arydshln}

\usepackage[pagebackref,breaklinks,colorlinks]{hyperref}

\title{\LARGE \bf VRS-NeRF: Visual Relocalization with Sparse Neural Radiance Field}

\author{Fei Xue$^1$, Ignas Budvytis$^1$, Daniel Olmeda Reino$^2$, Roberto Cipolla$^1$ \\
	$^1$University of Cambridge,  $^2$Toyota Motor Europe\\
	\{fx221,ib255,rc10001\}@cam.ac.uk,  daniel.olmeda.reino@toyota-europe.com
}

\begin{document}

\maketitle
\thispagestyle{empty}
\pagestyle{empty}


\input{sections/abstract}
\input{sections/introduction}

\input{sections/relatedwork}
\input{sections/method}

\input{sections/experiment}

\input{sections/conclusion}



\bibliographystyle{IEEEtran}
\bibliography{IEEEexample}

\end{document}

%% file: sections/abstract.tex
\begin{abstract}
	Visual relocalization is a key technique to autonomous driving, robotics, and virtual/augmented reality. After decades of explorations, absolute pose regression (APR), scene coordinate regression (SCR), and hierarchical methods (HMs) have become the most popular frameworks. However, in spite of high efficiency, APRs and SCRs have limited accuracy especially in large-scale outdoor scenes; HMs are accurate but need to store a large number of 2D descriptors for matching, resulting in poor efficiency. In this paper, we propose an efficient and accurate framework, called VRS-NeRF, for visual relocalization with sparse neural radiance field. Precisely, we introduce an explicit geometric map (EGM) for 3D map representation and an implicit learning map (ILM) for sparse patches rendering. In this localization process, EGP provides priors of spare 2D points and ILM utilizes these sparse points to render patches with sparse NeRFs for matching. This allows us to discard a large number of 2D descriptors so as to reduce the map size. Moreover, rendering patches only for useful points rather than all pixels in the whole image reduces the rendering time significantly. This framework inherits the accuracy of HMs and discards their low efficiency. Experiments on 7Scenes, CambridgeLandmarks, and Aachen datasets show that our method gives much better accuracy than APRs and SCRs, and close performance to HMs but is much more efficient.  Source code is available at \url{https://github.com/feixue94/vrs-nerf}.
\end{abstract}

%% file: sections/introduction.tex
\section{INTRODUCTION}
\label{sec:introduction}
Visual localization aims to estimate the rotation and position of a given image captured in a known environment. As a fundamental computer vision task, visual localization is the key technique to various applications such as virtual/augmented reality (VR/AR), robotics, and autonomous driving. After several decades of exploration, many excellent methods have been proposed~\cite{lbr, smc, hfnet, posenet} and can be roughly categorized as absolute pose regression (APR)~\cite{posenet,posenet-geo2017,lsg,glnet,atloc2020,mapnet}, scene coordinate regression (SCR)~\cite{scr2013,dsac,sc-wls2022,ace2023}, and hierarchical methods (HM)~\cite{as,hfnet,lbr}. APRs embed the map into high-level pose features and predict the 6-DoF pose with multi-layer perceptions (MLP); they are fast especially for large-scale scenes, but have limited accuracy due to implicit 3D information representation. Different with APRs, SCRs regress 3D coordinates for pixels to build 2D-3D matches directly and estimate the pose with PnP~\cite{epnp} and RANSAC~\cite{ransac1981}. Despite the high accuracy in indoor environments, SCRs can't scale up to outdoor large-scale scenes. Instead of using an end-to-end 2D-3D matches prediction, HMs adopt global features~\cite{netvlad,patchnetvlad2021,gem2018} to search reference images in the database and then build correspondences between keypoints extracted query and reference images; these 2D-2D matches are lifted to 2D-3D matches and used for absolute pose estimation with PnP~\cite{epnp} and RANSAC~\cite{ransac1981} as SCRs. Because of high accuracy and flexibility, HMs are widely used recently. However, the huge memory cost of 2D keypoints storage impairs their efficiency in real applications. 


\begin{figure}[t]
	\centering
	\includegraphics[width=1.\linewidth]{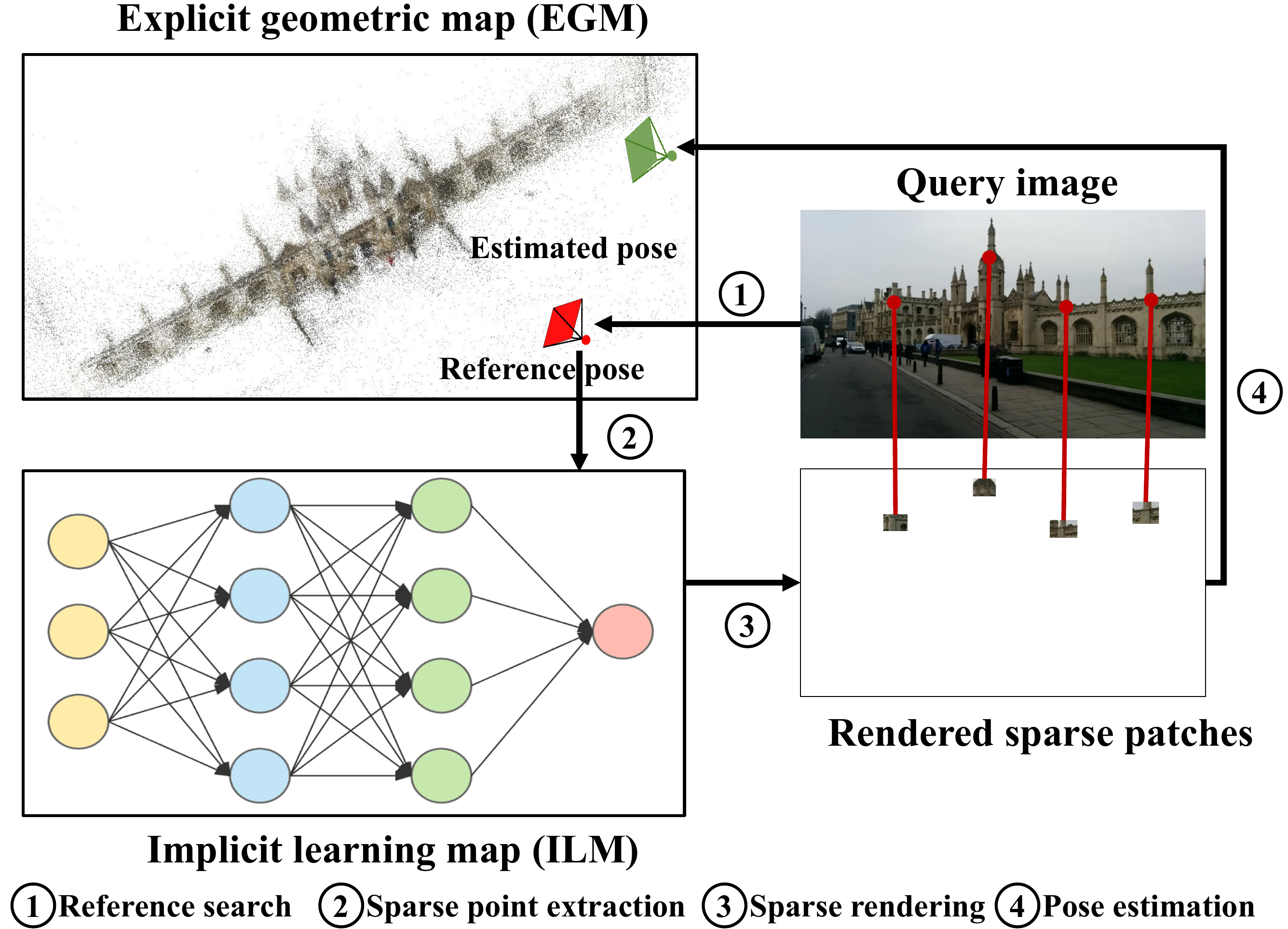}
	\caption{\textbf{Overview of our framework}. We use an explicit geometric map (EGM) and an implicit learning map (ILM) to represent the environment. At test time, we first use the query image to find a reference image in EGM , then the pose and sparse points of this reference image are fed into ILM for sparse patch rendering. Finally the matches between the query and rendered patches are used for pose estimation with PnP~\cite{epnp} and RANSAC~\cite{ransac1981}. }
	\label{fig:overview}
\end{figure}

In this paper, we aim to find an efficient and accurate solution to large-scale visual localization task. To achieve this, we seek help from neural radiance fields (NeRFs). NeRFs are first proposed for view synthesis~\cite{nerfs2021}. Due to their powerful scene and object representation ability, NeRFs have been widely used for many other tasks including visual localization~\cite{nerfloc2023, lens2022}. Although LENS~\cite{lens2022} and NeRF-loc~\cite{nerfloc2023} have applied NERFs to APRs and SCRs respectively, their performance especially in outdoor scenes is still limited. Besides, direct usage of NeRFs for localization is inefficient as rendering all pixels of an image is slow.

Instead, we adopt a hybrid map of using NeRFs for efficient localization by rendering only useful sparse pixels. The hybrid map consists of two parts: explicit geometric map (EGM) and implicit learning map (ILM). EGM contains the sparse 3D points along with their 2D observations on reference images. ILM is the implicit map represented by NeRFs. At test time, 2D observations of reference images provide prior sparse pixel locations and camera poses as input to NeRFs. NeRFs return RGB values of each sparse pixel. In order to improve the accuracy, we render a patch with constant size for each pixel. These rendered patches are further used to build 2D-3D matches for absolute pose estimation with PnP~\cite{epnp} and RANSAC~\cite{ransac1981}.

An overview of our framework is shown in Fig.~\ref{fig:overview}. With EGM and ILM, our method is able to render useful pixels online as opposed to relying on offline 2D descriptors for matching, making the localization system much more efficient. To allow current NeRFs work in large-scale scenes, we adopt a clustering-based strategy to divide the scene into smaller ones adaptively and automatically. The contributions of our method are summarized as follows:

\begin{itemize}
	\item We propose a hybrid method combining explicit geometric map and implicit learning map for visual localization, making localization system efficient and accurate.
	\item Instead of rendering images, we render patches for useful sparse keypoints only, avoiding the time-consuming rendering process.
	\item We adopt a clustering-based strategy for scene division, enabling NeRFs to work in large-scale outdoor environments.
\end{itemize}

 Our experiments on the popular indoor 7Scenes~\cite{sevenscenes2013} and outdoor CambridgeLandmarks~\cite{posenet} and Aachen~\cite{aachen} datasets demonstrate that our method requires much less memory cost while preserving the accuracy. We hope this method could be a new baseline of applying NeRFs to visual localization task. We organize the rest of the paper as follows. In Sec.~\ref{sec:relatedwork}, we discuss related works. In Sec.~\ref{sec:method}, we introduce our method in detail. We test our approach in Sec.~\ref{sec:experiment} and conclude the paper in Sec.~\ref{sec:conclusion}.

%% file: sections/relatedwork.tex
\section{Related Work}
\label{sec:relatedwork}

In this section, we discuss related works about visual localization and NeRFs.

\textbf{Visual localization.} Visual localization methods can be roughly categorized as absolute pose regression (APR), scene coordinate regression (SCR), and hierarchical methods (HM). Posenet~\cite{posenet} is the first work introducing APR. Due to its simplicity, high memory and time efficiency, especially in large-scale scenes, a lot of variants have been proposed by introducing geometric loss~\cite{posenet-geo2017}, multi-view constraints~\cite{mapnet,lsg,glnet,pogonet2021,gtcar2022,vcr2021}, view synthesis~\cite{dfanet2019}, feature selection~\cite{atloc2020,gtcar2022} and additional training data generation~\cite{lens2022}. However, the accuracy is still limited because of the retrieval nature of APRs~\cite{sattler2019understanding}.

SCRs first regress 3D coordinates for each pixel in the query image and then estimate the pose with PnP~\cite{epnp} and RANSAC~\cite{ransac1981}. Initially, this is achieved via random forest with RGBD as input~\cite{scr2013}. Later on, DSAC~\cite{dsac} and its variants~\cite{dsac*,dsac++} extend it to RGB input and replace random forest with CNNs. More recently, hierarchical prediction~\cite{hscnet2020}, semantic-aware prediction~\cite{localinstance}, and the separation of backbone and prediction head~\cite{ace2023}, to name a few, are introduced for better accuracy and training efficiency. SCRs compress the map into a compact network and give very accurate poses in small scenes such as indoor environments~\cite{sevenscenes2013,twelvescenes2016}, but have limited accuracy in large-scale scenes including CambridgeLandmarks~\cite{posenet} and Aachen dataset~\cite{aachen}.

HMs~\cite{hfnet,lbr} estimate the pose of a query image by first finding reference images in the database, then building matches between keypoints in query and reference images, and finally compute the pose from associated 2D-3D matches lifted from 2D-2D ones with PnP~\cite{epnp} and RANSAC~\cite{ransac1981}. Traditionally, handcrafted SIFT~\cite{sift} or ORB~\cite{orb} features along with BoWs~\cite{bow2008,bow2012} are widely used for the first two steps~\cite{as}. Due to the sensitivity of handcrafted features to illumination and seasonal changes, learned local features~\cite{superpoint,r2d2,disk,sfd22023,lift,d2-net}, e.g., Lift~\cite{lift}, SuperPoint (SP)~\cite{superpoint}, SFD2~\cite{sfd22023} and global features, e.g., NetVLAD (NV)~\cite{netvlad}, GeM~~\cite{gem2018} are used to replace classic ones. To further improve the accuracy, graph-based matchers, e.g., SuperGlue (SG)~\cite{superglue}, SGMNet~\cite{sgmnet}, IMP~\cite{imp2023} are proposed to replace nearest matching for better 2D-2D matches. Nowadays, the combination of SPP+NV+SG~\cite{superpoint,netvlad,superglue} has set the new state-of-art on public datasets~\cite{sevenscenes2013,twelvescenes2016,posenet,aachen}. Despite the outstanding accuracy, HMs need to store a large amount of local features to build 2D-3D correspondences, impairing the memory efficiency. 

Our method absorbs advantages of HMs in terms of accuracy by introducing an explicit geometry map (EGM) to preserve geometric information. Additionally, we leverage the implicit learning map (ILM) for rendering useful pixels for high efficiency instead of storing redundant 2D descriptors for higher efficiency.

\textbf{Neural radiance fields (NeRFs).} NeRFs are first introduced for view synthesis~\cite{nerfs2021}. After two-year exploration, many excellent works~\cite{mipnerf2022,zipnerf2023,ing2022,gaussiansplatting2023} have been proposed to achieve higher quality and faster speed. Although some of them~\cite{gaussiansplatting2023} report real-time rendering performance, they require to storage additional intermediate features, resulting in heavy memory cost. As NeRFs have impressive ability of scene representation, some works have applied them to visual localization task. LENS~\cite{lens2022} uses NeRFs to render more images from different viewpoints for training absolute pose regression and obtains better accuracy. NeRF-loc~\cite{nerfloc2023} adopts a conditional NeRF to render 3D descriptors for matching. Despite its promising performance, it needs support images, features and depth maps as input, which leads to low memory efficiency. Moreover, the dense rendering of all pixels further impairs its time efficiency.

Different with LENS~\cite{lens2022} and NeRF-loc~\cite{nerfloc2023}, we use a hybrid map consisting of EGP and ILP for online sparse rendering to reduce time and memory cost and yield higher accuracy as well.

%% file: sections/method.tex
\section{Method}
\label{sec:method}

\begin{figure*}[t]
	\centering
	\includegraphics[width=.95\linewidth]{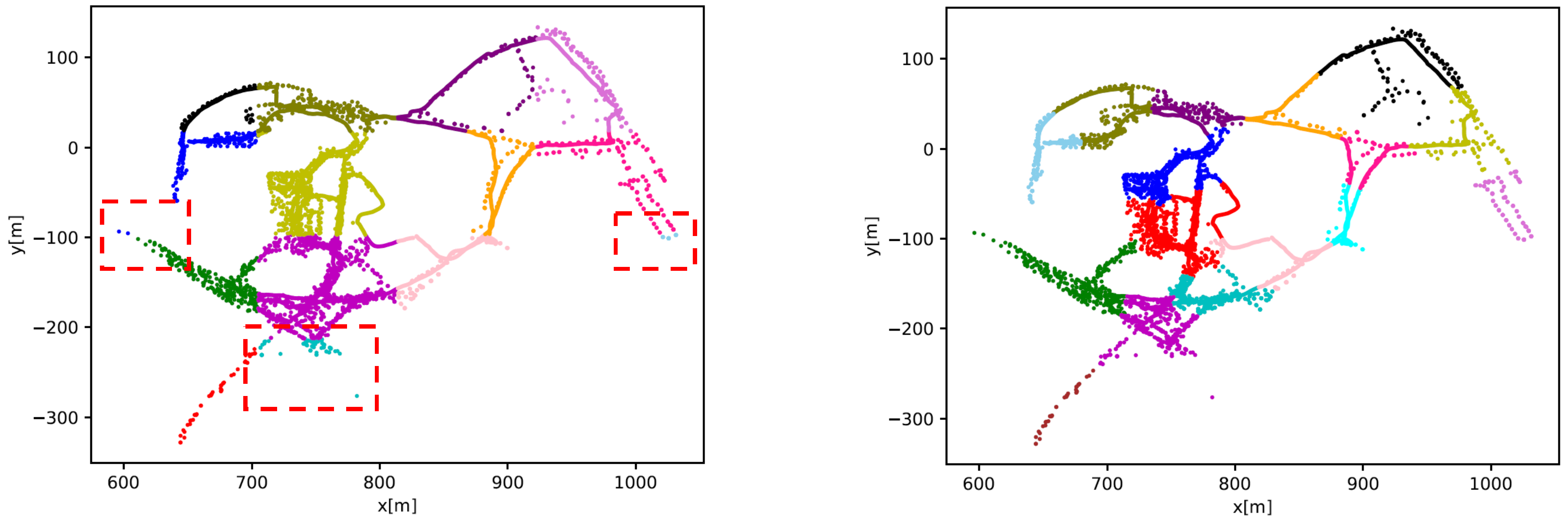}
	\caption{\textbf{Visualization of scene division on Aachen dataset~\cite{aachen}}. The uniform division of scene leads to imbalanced pieces (left) and our clustering on reference poses gives more balanced results (right).}
	\label{fig:aachen_clu}
\end{figure*}

In this section, we first introduce the explicit geometric map and implicit learning map in Sec.~\ref{sec:method:egm} and Sec.~\ref{sec:method:ilm}, respectively. Then, we describe the process of localization with sparse NeRFs in Sec.~\ref{sec:method:localization}.

\subsection{Explicit Geometric Map}
\label{sec:method:egm}

We first build the explicit geometric map (EGM). Given reference images, we adopt current state-of-the-art SfM library colmap~\cite{colmap2016} to reconstruct the 3D environment. EGM consists of a set of 3D points $\mathcal{X}=\{X_1, ..., X_m\}$ and reference images $\mathcal{I}=\{I_1, ..., I_n\}$. Each reference image $I_i$ has several 2D keypoints $\mathcal{P}_i=\{P^{i}_1, ..., P^{i}_k\}$ corresponding to 3D points in $\mathcal{X}$. Each reference image $I_i$ also has a global descriptor $\mathbf{v}_i$ provided by NetVLAD~\cite{netvlad} to search similar reference images for a given query image $I_q$. It is worthy to note that in EGM only global descriptors are retained and local descriptors of 2D keypoints are discarded due to their huge memory cost.

\subsection{Implicit Learning Map}
\label{sec:method:ilm}

The implicit learning map (ILM) is a NeRF-based model compressing the scene in a single network implicitly. NeRF renders the RGB values of a single ray $\mathbf{r}(t)=\mathbf{o} + t\mathbf{d}$ where $\mathbf{o}$ and $\mathbf{d}$ are the origin and direction of the ray, and $t$ is distance along the ray. Usually, the ray is split into a set of intervals $T_i=[t_i, t_{i+1})$. The features generated for $T_i$ with function $\gamma$ are fed into a MLP for the density $\tau$ and color $\mathbf{c}$ prediction:
\begin{align}
	\tau_i, &\mathbf{c}_i = MLP(\gamma(\mathbf{r}(T_i)), \mathbf{d};\Theta), \mathbf{C}(\mathbf{r}, \mathbf{t}) = \sum_{i}^{}\omega_i \mathbf{c}_i, \\	
	\omega_i &= (1 - e^{-\tau_i(t_{i+1} - t_i)})e^{-\sum_{i' < i} \tau_{i'}(t_{i'+1} - t^{i'})}.
\end{align}
$\mathbf{C}(\mathbf{r}, \mathbf{t})$ is the predicted pixel value. For better performance, it is optimized from coarse to fine as:
\begin{align}
	\sum_{\mathbf{r} \in \mathcal{R}}0.1L_{rec}(\mathbf{C}(\mathbf{r},\mathbf{t}^c), \mathbf{C}^{*}(\mathbf{r}))+L_{rec}(\mathbf{C}(\mathbf{r},\mathbf{t}^{f}), \mathbf{C}^{*}(\mathbf{r})).
\end{align}
$\mathcal{R}$ is the set of rays for training. $\mathbf{t}^{c}$ and $\mathbf{t}^{f}$ are predicted coarse and fine distances. $\mathbf{C}^{*}(\mathbf{r})$ is the ground-truth color. $L_{rec}$ is the mean squared error. We adopt Mip-NeRF~\cite{zipnerf2023} as the base model in our ILM.

\textbf{Scene division.} A single NeRF may not be able to represent a scene of large scale~\cite{blocknerf2022}, so we need to divide the scene into several pieces. A straightforward method is to divide the scene uniformly. However, this usually makes sub-scenes imbalance because different areas with different density of objects. Instead, we propose to leverage to priors of reference images which reconstruct the scene. In detail, we perform clustering on the poses of reference images with K-means~\cite{kmeans++2007} with a given number of clusters $N$. This strategy allows us to segment the scene into more balanced pieces, as shown in Fig.~\ref{fig:aachen_clu}.

In our experiment, we divide the large-scale Aachen dataset~\cite{aachen} into 16 sub-scenes and train 16 separate NeRFs $\Theta_1, ..., \Theta_{16}$ independently to represent each of them, respectively.

%

\input{tables/table_7scenes}

\subsection{Localization with Sparse NeRFs}
\label{sec:method:localization}

Given a query image $I_q$, we first extract global descriptor $\mathbf{v}_q$ and a set of 2D local keypoints $\mathcal{P}^{q}=\{P^{q}_1, ..., P^{q}_k\}$.

\textbf{Reference image search.} As EGM contains global descriptors of all reference images, we perform nearest matching to find top $k$ candidate images with the smallest distances between reference and query global descriptors. These candidate reference images are sorted as $\mathcal{I}^c=\{I^{c}_1, ..., I^{c}_k\}$ in ascending order according to their distances to $v_q$.

\textbf{Matching with online renderings.} As~\cite{hfnet}, we perform 2D-2D matching between the query and all candidate reference images independently. Since the EGM does not contain any 2D descriptors, we render the patches with ILM on the fly. For a keypoint $P^{c}_{ij}$ from candidate reference image $I^{c}_i$, we generate $\lambda_r \times \lambda_r$ rays at center of $P^{c}_{ij}$ denoted as $\mathcal{R}^{c}_{ij}$. $\mathcal{R}^{c}_{ij}$ along with the pose of $I^{c}_i$ is then fed into ILM for RGB values prediction. $\mathcal{R}^{c}_{ij}$ is reshaped as a patch $p^{c}_{ij}$ with size of $\lambda_r \times \lambda_r$. $p^{c}_{ij}$ is finally used as the input to a local feature network~\cite{superpoint,sfd22023} as obtain the descriptor $d^{c}_{ij}$ and score $s^{c}_{ij}$ for 2D-2D matching. The whole process can be described as:
\begin{align}
	\mathcal{R}^{c}_{ij} &= f_r(P^{c}_{ij}, \lambda_r), 
	\mathcal{C}^{c}_{ij} = f_{nerf}(\mathcal{R}^{c}_{ij}, \Theta), \\
	p^{c}_{ij} & = f_p(\mathcal{C}^{c}_{ij}), 
	(d^{c}_{ij}, s^{c}_{ij}) = f_{l}(p^{c}_{ij}),
\end{align} 
where $f_r, f_{nerf}, f_p, f_l$ are the ray generation, NeRF with weights of $\Theta$, patch, and local feature extraction functions.

Compared with image-wise rendering, our sparse rendering focuses only on the useful keypoints with corresponding 3D points in the map, reducing the time significantly. For example, an image with size of $480\times640$ from 7Scenes dataset~\cite{sevenscenes2013} has 307,200 rays in total. However, for 500 keypoints with patch size of $15\times15$, the number of rays is 112,500 which is 2.7$\times$ fewer. As patches have overlap, for each unique ray, we only render it once, which further makes the rendering efficient.  The online rendering of patches introduce additional time cost of feature extraction, because they are only patches with limited size, the additional time cost is limited.

\textbf{Pose estimation.} The 2D-2D matches between the query $I^{q}$ and candidate reference images $\mathcal{I}^{c}$ are lifted to 2D-3D matches between $I^{q}$ and EGM. Then, we use RANSAC~\cite{ransac1981} and EPnP~\cite{epnp} to recover the absolute pose.

%% file: tables/table_7scenes.tex
\setlength{\tabcolsep}{8pt}

\begin{table*}[t]
	\centering
		\begin{tabular}{llcccccccc}
			\toprule
			 Group & Method & Chess & Fire &  Heads & Office & Pumpkin & Kitchen & Stairs & Average (\%)\\
			 
			 \midrule
			\multirow{8}{*}{APRs}  
			& Posenet~\cite{posenet} & 13, 4.5 & 27, 11.3 & 17, 13.0 & 19, 5.6 & 26, 4.8 & 23, 5.4 & 35, 12.4 & -\\
			& MapNet~\cite{mapnet} & 8, 3.3 & 27, 11.7 & 18, 13.3 & 17, 5.2 & 22, 4.0 & 23, 4.9 & 30, 12.1 & -\\
			& LsG~\cite{lsg} & 9, 3.3 &  26, 10.9 &  17, 12.7 & 18, 5.5 & 20, 3.7 & 23, 4.9 & 23, 11.3 & - \\
			& AtLoc~\cite{atloc2020} &  10, 4.1 & 25, 11.4 & 16, 11.8 & 17, 5.3 & 21, 4.4 & 23, 5.4 & 26, 10.5 & -\\
			& GLNet~\cite{glnet} & 8, 2.8 & 26, 8.9 & 17, 11.4 & 18, 5.1 & 15, 2.8 & 25, 4.5 & 23, 8.8 & -\\
			& SC-wLS~\cite{sc-wls2022} & 3, 0.8 & 5, 1.1 & 3, 1.9 & 6, 0.9 & 8, 1.3 & 9, 1.4 & 12, 2.8 & -\\
			& PAEs~\cite{pae2022} & 12, 5.0 & 24, 9.3 & 14, 12.5 &  19, 5.8 &  18, 4.9 &  18, 6.2 & 25, 8.7 & -\\
			& LENS~\cite{lens2022} & 3, 1.3 & 10, 3.7 & 7, 5.8 & 7, 1.9 & 8, 2.2& 9, 2.2 & 14, 3.6 & -\\
			\midrule
			\multirow{3}{*}{SCRs}
			& DSAC* & 2, 1.1 & 2, 1.2 & 1, 1.8 & 3, 1.2 & 4, 1.3 & 4, 1.7 & 3, 1.2 & 96.0 \\
			& ACE~\cite{ace2023} & 2, 1.1 & 2, 1.8 & 2, 1.1 & 3, 1.4 & 3, 1.3 & 3, 1.3 & 3, 1.2 & 97.1\\
			& NeRF-loc~\cite{nerfloc2023} & 2, 1.1 & 2, 1.1 & 1, 1.9 & 2, 1.1 & 3, 1.3 & 3, 1.5 & 3, 1.3 & 89.5 \\
			\midrule
			\multirow{2}{*}{HMs}
			& SP+SG~\cite{superpoint,superglue} & 0, 0.1 & 1, 0.2 & 0, 0.2 & 1, 0.2 & 1, 0.1 & 0, 0.1 & 2, 0.6 & 95.7\\
			& SFD2+IMP~\cite{sfd22023,imp2023} & 0, 0.1 & 1, 0.2 & 0, 0.2 & 1, 0.2 & 1, 0.2 & 0, 0. & 2, 0.5 & 95.7\\
			\midrule
			& \textbf{Ours} & 0, 0.1 & 1, 0.2 & 0, 0.2 & 1, 0.2 & 1, 0.2 & 0, 0.1  & 3, 0.8 & 93.1 \\
			
			\bottomrule
		\end{tabular}
\caption{\textbf{Localization accuracy on 7Scene dataset~\cite{sevenscenes2013}.} We report the median translation (cm) and rotation ($^\circ$) errors and the average success ratio of poses within error of $5cm, 5^\circ$.}
\label{tab:7scenes}
\end{table*}

%% file: sections/experiment.tex
\section{Experiment}
\label{sec:experiment}

In this section, we compare the performance of our method with previous approaches for localization in terms both accuracy and efficiency.

\input{tables/table_cambridgelandmarks}

\subsection{Implementation}
\label{sec:exp:imp}
We implement our model on PyTorch~\cite{pytorch}. We use Mip-NeRF~\cite{mipnerf2022} as our basic NeRF model for scene representation. Each model is trained with Adam optimizer~\cite{adam}, batch size of 1024 on a single RTX 3090 GPU for 50,000 iterations in total. The initial leaning rate is set to 0.001 and is decayed at the ratio of 1e-8 after 5,000 iterations. The patch size $\lambda_r$ is set to 15 in our experiments.

\subsection{Datasets, Metrics and Baselines}
\label{sec:exp:dataset}

\textbf{Dataset.} We evaluate our method on three public datasets including the indoor 7Scenes~\cite{sevenscenes2013} and outdoor CambridgeLandmarks~\cite{posenet} and Aachen~\cite{aachen} datasets. For 7Scenes and CambridgeLandmarks, we assign each sub-scene a NeRF model. As Aachen dataset is a large-scale outdoor scene, we divide it into 16 sub-scenes with the clustering strategy as introduced in Sec.~\ref{sec:method:ilm}. Following previous methods~\cite{hfnet,lbr,sfd22023}, NetVLAD (NV)~\cite{netvlad} is adopted to provide 20, 20 and 50 candidate reference images for query images in 7Scenes, CambridgeLandmarks, and Aachen datasets, respectively. We use SFD2~\cite{sfd22023} and IMP~\cite{imp2023} as the local feature and matcher because SFD2+IMP gives better performance than SP+SG on Aachen dataset and has smaller map size.

\textbf{Metrics.} As~\cite{dsac*,hfnet,as}, we report the median rotation and translation errors on 7Scenes~\cite{sevenscenes2013} and CambridgeLandmarks~\cite{posenet}. Besides, the success ratio of pose error within $5cm,5^\circ$ and $25cm, 5^\circ$ are also provided. For Aachen dataset~\cite{aachen}, we use the official metric of success ratio at error thresholds of $0.25m,2^\circ$, $0.5m,5^\circ$ and $5m,10^\circ$. 

By introducing sparse NeRFs, we aim to reduce the high memory cost of HMs and the time cost of image-wise rendering. Therefore, we additionally analyze the memory and time efficiency.

\textbf{Baselines.} We compare our system with previous APRs~\cite{posenet,glnet,lsg,sc-wls2022,atloc2020,pae2022,mapnet}, SCRs~\cite{dsac,dsac*,ace2023} and hierarchical methods~\cite{as,hfnet,lbr,sfd22023}. Besides, we also compare our approach with previous NeRF-based methods including LENS~~\cite{lens2022} and NeRF-loc~\cite{nerfloc2023}.

\subsection{Pose Accuracy Analysis}
\label{sec:exp:pose}

\textbf{7Scenes.} As shown in Table~\ref{tab:7scenes}, we compare our approach with previous APRs~\cite{posenet,mapnet,lsg,glnet,atloc2020,sc-wls2022,pae2022,lens2022}, SCRs~\cite{dsac*,ace2023,nerfloc2023} and HMs~\cite{superpoint,superglue,sfd22023,imp2023}. APRs give the largest errors because their similar behaviors to image retrieval in the localization process~\cite{sattler2019understanding}, resulting in limited pose accuracy. Since most APRs only report median errors, their success ratios are not available. SCRs obtain much higher accuracy than APRs due to their explicit 3D coordinates regression. HMs achieve the best accuracy in terms of median errors. However, they are less robust to textureless areas due to the reliance on sparse keypoints, so they report slightly worse accuracy ratio than some SCRs such as DSAC*~\cite{dsac*} and ACE~\cite{ace2023}. Although our approach renders sparse patches for localization, it yields close performance to HMs and outperforms APRs and SCRs significantly in terms median errors. Similar to HMs, our model is also sensitive to textureless regions. As our EGM inherits advantages of HMs, it outperforms previous approachs LENS~\cite{lens2022} and NeRF-loc~\cite{nerfloc2023} which introduce NeRFs to APRs and SCRs, respectively.

\textbf{CambridgeLandmarks.} Table~\ref{tab:cambridgelandmarks} demonstrates the results of previous and our methods on the CambridgeLandmarks dataset~\cite{posenet}. We report the median translation (cm) and rotation ($^\circ$) errors and the success ratio of poses within error threshold of $25cm, 2^\circ$. Similar to results on 7Scenes dataset~\cite{sevenscenes2013}, APRs give over 2$\times$ larger errors than SCRs due to the missing 3D information embedding. 

SCRs report promising accuracy in terms of median translation and rotation errors. However, their success ratios within the error threshold of $25cm, 2^\circ$ are much worse than HMs. Even the state-of-the-art DSAC*~\cite{dsac*} and ACE~\cite{ace2023} fail to achieve comparable accuracy to HMs. These comparisons reveal that SCRs are not that accurate as they are expected to be in outdoor scenes. Note that each sub-scene in CambridgeLandmarks dataset such as Kings College is of size about $500m\times50m$, far from large-scale environments. We hope the future methods especially SCRs could also include the success ratio metric so as to provide a deeper understanding of their performance.

HMs are still the most accurate methods in terms of both median errors and success ratios. As our approach also preserves the explicit geometric information as explicit geometric map, its results are as accurate as HMs' and are much more accurate than APRs and SCRs.

Compared with prior NeRF-based LENS~\cite{lens2022} and NeRF-loc~\cite{nerfloc2023}, our system also achieves the significantly better accuracy.

\input{tables/table_aachen}

\textbf{Aachen.} We finally show the results on Aachen dataset~\cite{aachen}. As Aachen dataset is a large-scale dataset consisting of images with illumination, season and large-viewpoint changes, few APRs and SCRs report numbers on this dataset. We follow the official metric by showing the success ratio of poses at error thresholds of $0.25m,2^\circ$, $0.5m,5^\circ$ and $5m,10^\circ$. SCRs including ESAC~\cite{esac2019} and HSCNet~\cite{hscnet2020} give relatively worse accuracy especially on night images due to severe illumination changes. Among HMs, the combination of SFD2 and IMP~\cite{sfd22023,imp2023} achieves the state-of-the-art performance.

Due to poor rendering quality of insufficient observations and large illumination changes, our method gives poor performance when the patch size is 15. When using larger patch size (31), our model gives promising accuracy. However, a large gap between the performance of our method and state-of-the-art SFD2+IMP. As our approach is the first the apply NeRFs on large-scale Aachen dataset, we hope more works can make it better in the future.

\begin{figure*}[t]
	\centering
	\includegraphics[width=1.\linewidth]{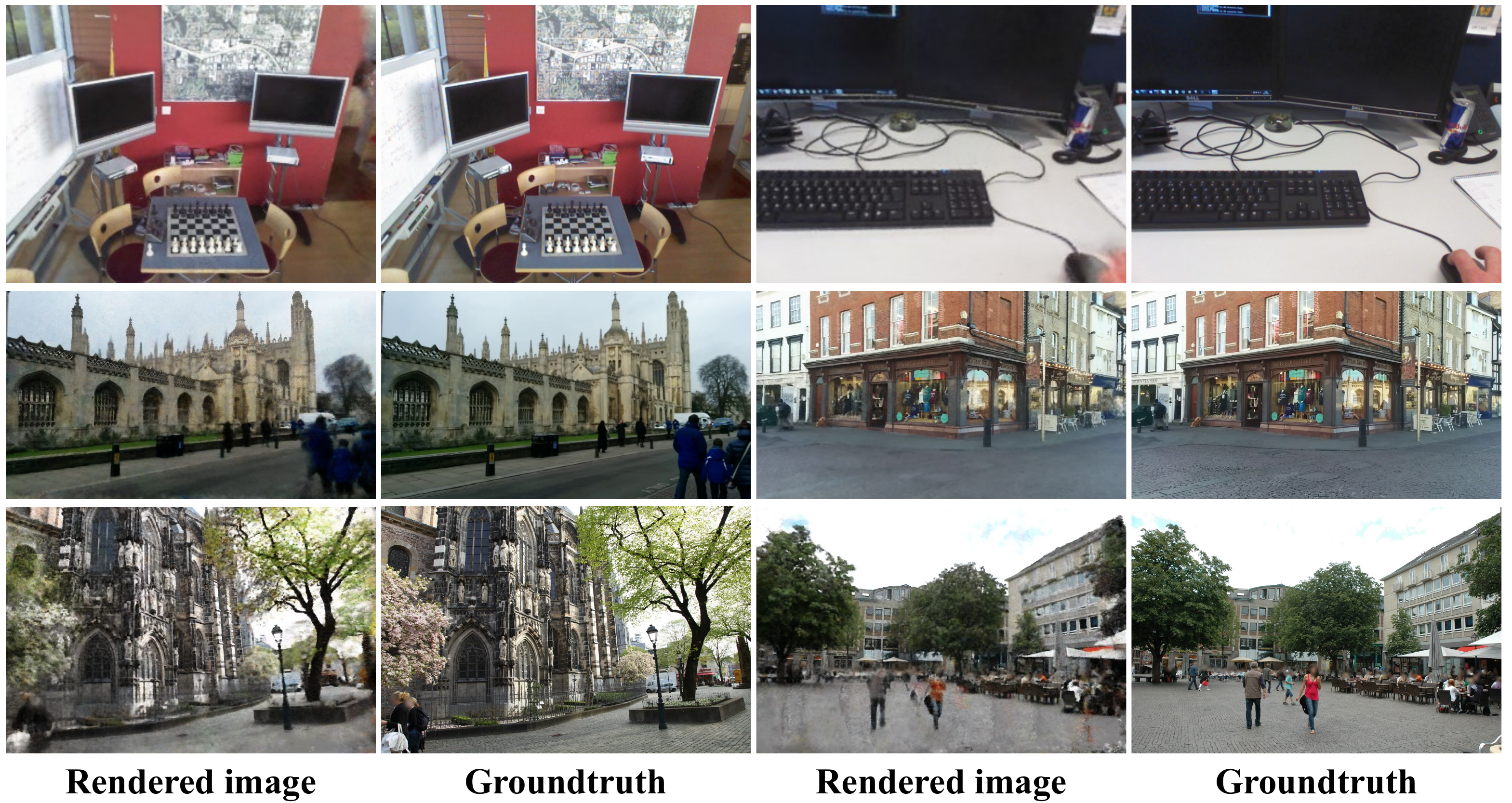}
	\caption{\textbf{Visualization of rendered image.} We visualize the rendered and groundtruth images from 7Scenes~\cite{sevenscenes2013} (top), CambridgeLandmarks~\cite{posenet} (middle) and Aachen~\cite{aachen} (bottom) datasets.}
	\label{fig:rendered_image}
\end{figure*}

\begin{figure*}[t]
	\centering
	\includegraphics[width=1.\linewidth]{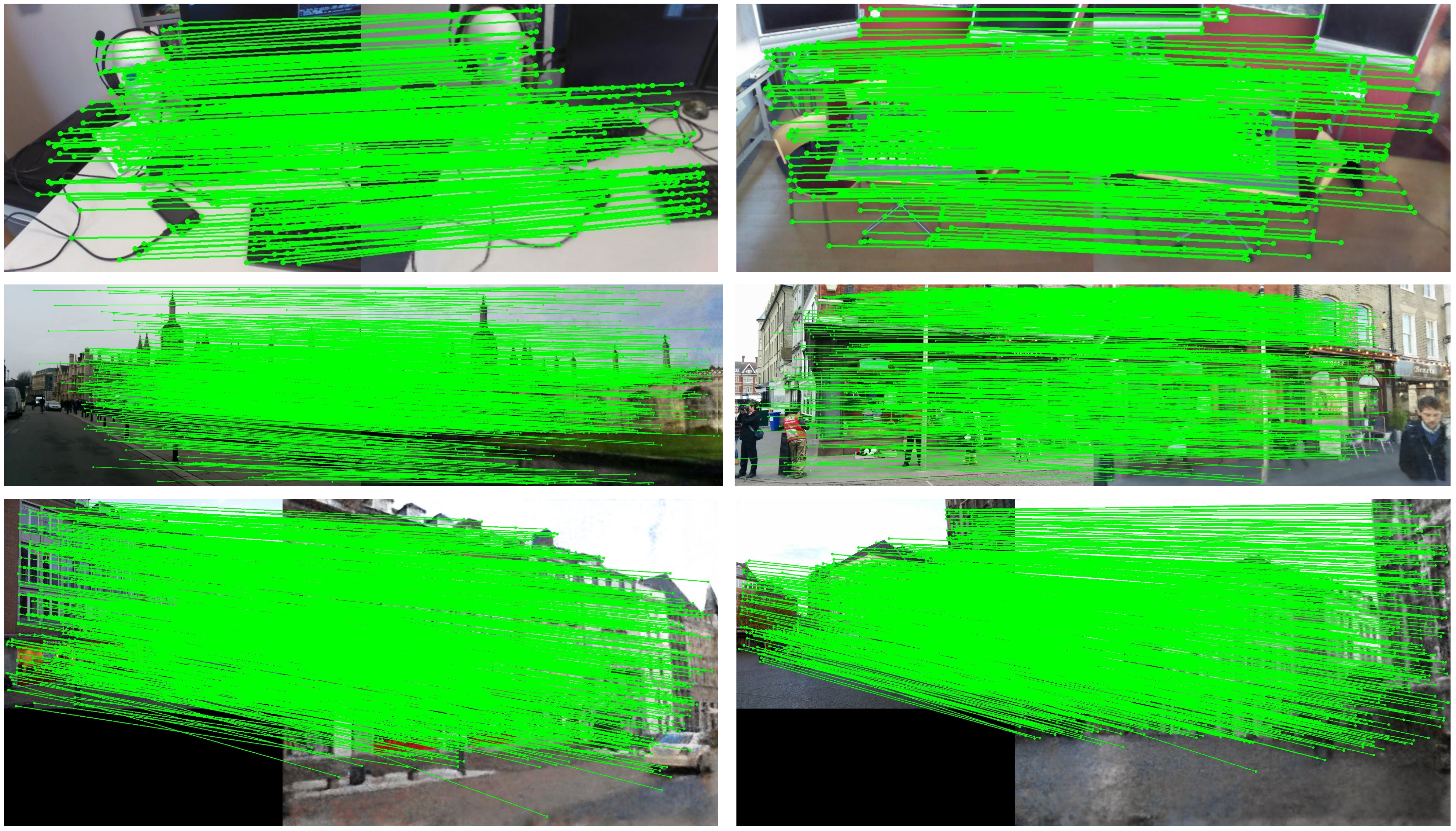}
	\caption{\textbf{Visualization of matches.} We visualize the matches between query (left) and reference (right) images from 7Scenes~\cite{sevenscenes2013} (top),CambridgeLandmarks~\cite{posenet} (middle) and Aachen~\cite{aachen} (bottom) datasets. Note that we perform matching the patches with size of $15\times15$ and show the whole rendered reference images for visualization only.}
	\label{fig:matching}
\end{figure*}

\subsection{Map size and Time Analysis}
\label{sec:exp:memory}

In this section, we analyze the map size and running time.

\input{tables/table_memory}

\textbf{Map size.} In table~\ref{tab:map_size}, we show the map size of APRs, SCRs, HMs and our method. For APRs and SCRs, the map size is the model size. For HMs, the map size is the sum of local descriptors, global descriptors, and 3D points. As our method discards local descriptors and introduces NeRFs, the map size of our model is the sum of global descriptors, 3D points and NeRFs. For simplicity, we report the numbers on a sub-scene of 7Scenes and CambridgeLandmarks and Aachen dataset.

Table~\ref{tab:map_size} shows that both APRs~\cite{posenet} and SCRs~\cite{dsac*,ace2023} are memory-efficient as they compress the map into neural networks at the cost of accuracy loss. HMs have a larger size of map due to the storage of 2D descriptors. SFD2+IMP~\cite{sfd22023,imp2023} has smaller size than SP+SG~\cite{superpoint,superglue} because SFD2 has smaller dimension of 2D descriptors. By discarding 2D descriptors, our method reduces the map size significantly. Our method has larger size of map on Aachen dataset as we use 16 NeRFs to represent the whole scene.

Although our method has much smaller size than HMs, its size is still much larger than that of APRs and SCRs. Therefore, how to design an accurate and efficient localization system is still a challenging task and is worth of further exploration in the future. 

\subsection{Ablation study}

We conduct an ablation study to explore the influence of different patch sizes to pose accuracy. Table~\ref{tab:ablation} shows as the patch size increases from $8\times9$ to $15\times15$, the pose accuracy also increases. It is more obvious on Kings College as it is an outdoor scene and the query and reference images have larger viewpoint and illumination changes. However, for indoor heads, due to little changes between the query and reference images, the improvement of increasing patch size is not obvious.

Moreover, as the patch size increases, it takes longer to render a patch. Therefore, the final solution is the balance between accuracy and efficiency. For indoor scenes without large changes between query and reference images, we suggest using smaller patch sizes to high efficiency. For outdoor scenes with large viewpoint, illumination changes between the query and reference images, a larger patch size can bring better accuracy.

\input{tables/table_ablation}

\begin{figure}[t]
	\centering
	\includegraphics[width=1.\linewidth]{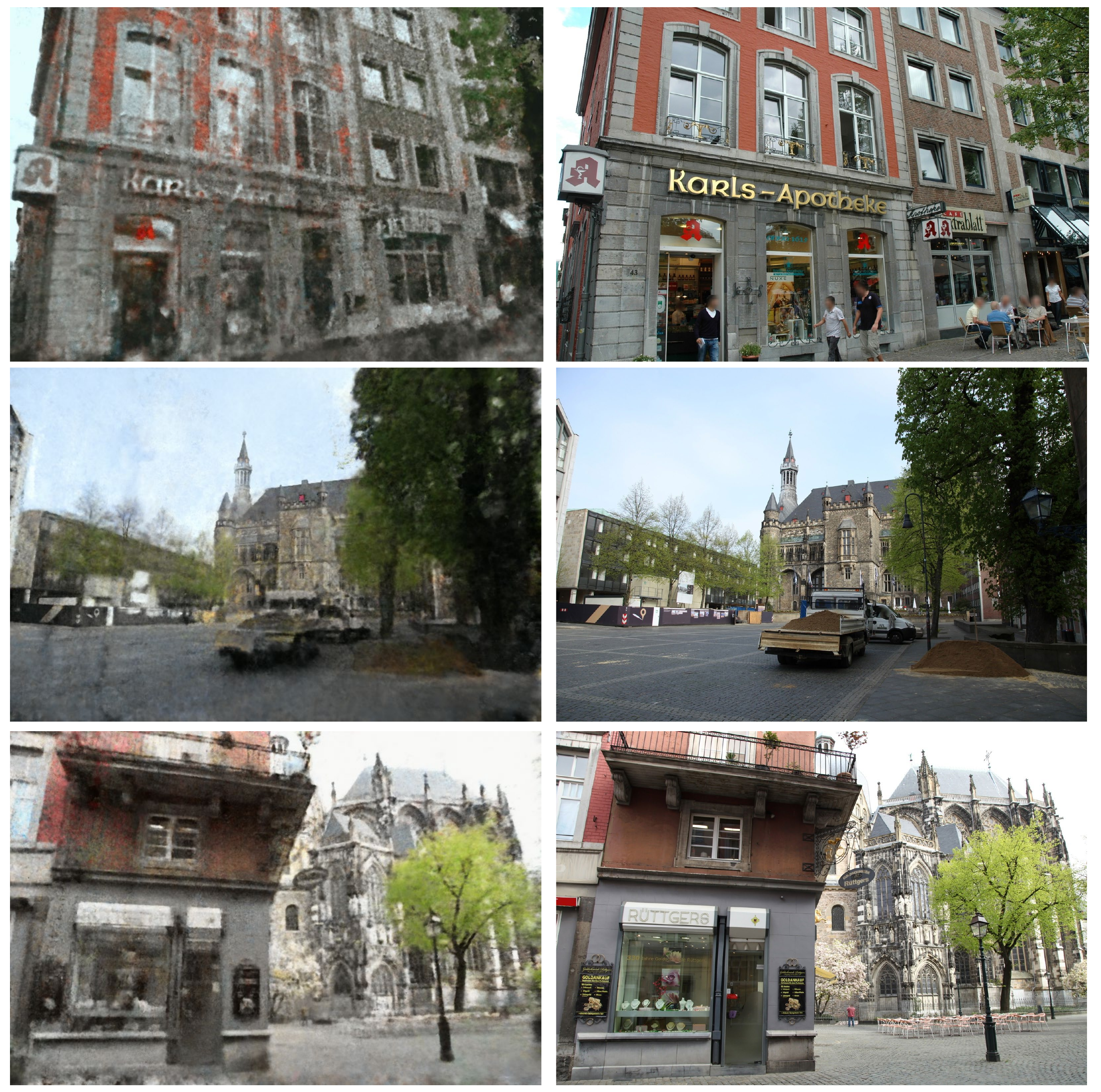}
	\caption{\textbf{Failed cases.}. We visualize the failed cases of rendered images (left) and their corresponding groundtruth images (right).}
	\label{fig:failed_case}
\end{figure}

\subsection{Qualitative results}

\textbf{Rendered image.} Fig.~\ref{fig:rendered_image} shows the rendered images in 7Scenes~\cite{sevenscenes2013}, CambridgeLandmarks~\cite{posenet} and Aachen~\cite{aachen} datasets. As 7Sevens dataset consists of only indoor scenes, the rendered images have higher quality. CambridgeLandmarks and Aachen datasets are both outdoor datasets, so there are still some artifacts on rendered images due to dynamic objects (e.g. pedestrian, trees), large depths and insufficient observations. Since our method relies mainly on sparse patches with corresponding 3D points the map, the influence of artifacts to localization is partially mitigated.

\textbf{Matching.} We also visualize the matches between query and rendered reference images in Fig.~\ref{fig:matching}. Although we show the whole image, in matching process, only patches with size of $15\times15$ at the center of sparse keypoints are used for matching. Due to the high quality of rendered images, we are able to obtain good matches between the query and rendered images, guaranteeing the success of localization even in outdoor scenes.

\textbf{Failed cases.} Fig.~\ref{fig:failed_case} shows failed cases of rendered images. We failed to render high quality images mainly because of insufficient 2D observations of physical 3D points and dynamic objects. The inaccurate camera poses also have negative influence to the rendering quality. This happens mainly on Aachen dataset probably because of insufficient images for some areas.


%% file: tables/table_cambridgelandmarks.tex
\setlength{\tabcolsep}{8.pt}

\begin{table*}[h]
	\centering
		\begin{tabular}{llcccccc}
			\toprule
			Group & Method & Kings College &  Great Court & Old Hospital & Shop Facade & St Mary Church & Average (\%)\\
			& & \multicolumn{6}{c}{$t (cm), R (^\circ), Percent (25cm, 2^\circ$)} \\
			\midrule
			\multirow{5}{*}{APRs} 
			&Posenet~\cite{posenet} & 88, 1.0 & 683, 3.5 & 88, 3.8 & 157, 3.3 & 320, 3.3 & - \\
			&MapNet~\cite{mapnet} & 107, 1.9 & 785, 3.8 & 149, 4.2 & 200, 4.5 & 194, 3.9 &  - \\
			&GLNet~\cite{glnet} & 59, 0.7 & 667, 3.0 & 50, 2.9 & 190, 3.3 & 188, 2.8 & - \\
			&PAEs~\cite{pae2022} & 90, 1.5 & - &  207, 2.6 & 99, 3.9 & 164, 4.2 &  \\
			&LENS~\cite{lens2022} & 33, 0.5 & - & 44, 0.9 & 27, 1.6 & 53, 1.6 & \\	
		
			\midrule
			\multirow{3}{*}{SCRs} 
			&DSAC*~\cite{dsac*} & 13, 0.4 & 40, 0.2 & 20, 0.3 & 6, 0.3 & 13, 0.4 & 64.88\\	
			&ACE~\cite{ace2023} & 28, 0.4  &  42, 0.2  & 31, 0.6 & 5, 0.3 & 19, 0.6 & 54.68 \\
			&NeRF-loc~\cite{nerfloc2023} & 7, 0.2 & 25, 0.1 &  18, 0.4 & 11, 0.2 &  4, 0.2 & - \\
			
			\midrule
			\multirow{2}{*}{HMs} 
			&SP+SG~\cite{superpoint,superglue} & 7, 0.1 & 12, 0.1 & 9, 0.2 & 2, 0.1& 4, 0.1 & 89.4\\  
			&SFD2+IMP~\cite{sfd22023,imp2023} & 7, 0.1 & 11, 0.1 & 10, 0.2 & 2, 0.1 & 4, 0.1 & 89.1\\
			\midrule
			&\textbf{Ours} & 9, 0.1 & - & 11, 0.2 & 2, 0.1 & 5, 0.2 & 89.3\\
			\bottomrule
		\end{tabular}
\caption{\textbf{Localization accuracy on CambridgeLandmarks dataset~\cite{posenet}.} We report the median translation (cm), rotation ($^\circ$) errors and the localization precision with translation and rotation errors within $25cm, 2^\circ$. $-$ indicates no values available. Due to poor image quality, we failed to train a NeRF model for Great Court. }
\label{tab:cambridgelandmarks}
\end{table*}

%% file: tables/table_aachen.tex
\setlength{\tabcolsep}{8.pt}

\begin{table}
	\scriptsize
	\centering
	\begin{tabular}{lcc}
		\toprule
		 Method & Day & Night\\
		& \multicolumn{2}{c}{$(2^\circ,0.25m)/ (5^\circ,0.5m)/ (10^\circ,5m)$} \\	
		\midrule
		ESAC~\cite{esac2019} & 42.6 / 59.6 / 75.5 & 3.1 / 9.2 / 11.2 \\
		HSCNet~\cite{hscnet2020} & 71.1 / 81.9 / 91.7 & 32.7 / 43.9 / 65.3 \\
		\midrule
		
		AS~\cite{as} & 85.3 / 92.2 / 97.9 & 39.8 / 49.0 / 64.3 \\
		LBR~\cite{lbr} & 88.3 / 95.6 / 98.8 & 84.7 / 93.9 / 100.0  \\
		SIFT ~\cite{sift} & 82.8 / 88.1 / 93.1 & 30.6 / 43.9 / 58.2 \\
		SP+SPG~\cite{superpoint, superglue} &89.6 / 95.4 / 98.8 & 86.7 / 93.9 / 100.0 \\
		SFD2+IMP~\cite{sfd22023,imp2023} & 89.7 / 96.5 / 98.9	& 84.7 / 94.9 / 100.0 \\
		\midrule
		\textbf{Ours (15)} & 	60.8 / 67.8 / 73.1 & 19.4 / 22.4 / 25.5 \\
		\textbf{Ours (31)} & 	70.1 / 76.9 / 80.9 &	44.9 / 51.0 / 62.2 \\
		\bottomrule			   
	\end{tabular}
	\caption{\textbf{Results on Aachen dataset~\cite{aachen,visbenchmark}.} We report the pose ratios within error thresholds of $2^\circ, 0.25m$, $5^\circ, 0.5m$ and $10^\circ, 5m$.}
	\label{tab:aachen}
\end{table}

%% file: tables/table_memory.tex
\setlength{\tabcolsep}{6.pt}

\begin{table}
	\scriptsize
	\centering
		\begin{tabular}{lccc}
			\toprule
			 Method & Chess~\cite{sevenscenes2013} & Kings College~\cite{posenet} & Aachen~\cite{aachen} \\
			 \midrule
			 Posenet~\cite{posenet} & 50MB & 50MB & -\\
			 \midrule
			 DSAC*~\cite{dsac*} & 28MB & 28MB & - \\
			 ACE~\cite{ace2023} & 4MB & 4MB & - \\
			 \midrule
			 SP+SG~\cite{superpoint,superglue} & 8.9GB & 13.3GB & 53.5GB\\
			 SFD2+IMP~\cite{sfd22023,imp2023} & 4.0GB & 6.6GB & 29.5GB \\
			 \midrule
			 \textbf{Ours} & 1.0GB & 0.76GB & 4.5GB\\
			\midrule
		\end{tabular}
\caption{\textbf{Map size.} We report the map size of PoseNet~\cite{posenet}, DSAC*~\cite{dsac*}, ACE~\cite{ace2023}, SP+SG~\cite{superpoint,superglue}, SFD2+IMP~\cite{sfd22023,imp2023} and our method.}
\label{tab:map_size}
\end{table}

%% file: tables/table_ablation.tex
\setlength{\tabcolsep}{4.pt}

\begin{table}
	\scriptsize
	\centering
	\begin{tabular}{lccc}
		\toprule
		Patch size & heads & Kings College & Rendering time (ms)  \\
		\midrule
		$9\times9$ & 0.3, 0.4, 97.2\% & 10.6, 0.2, 82.8\% & 0.65\\
		$11\times1$ & 0.2, 0.3, 97.8 & 9.5, 0.1, 88.9\% & 0.96\\	
		$13\times13$ & 0.2, 0.3, 98.6\% & 9.0, 0.1, 90.1\% & 1.35\\	
		$15\times15$ & 0.2, 0.3, 99.6\%& 9.1, 0.1, 90.7\% & 1.79\\ 
		$15\times15$ (GT) & 0.2, 0.3, 99.5\% & 8.2, 0.1, 94.2\% & 1.79\\
		\bottomrule
	\end{tabular}
	\caption{\textbf{Ablation study of patch size.} We verify the efficacy of different sizes of patches to the pose accuracy. For chess, we report the median translation (cm), rotation ($^\circ$) and pose accuracy at the error threshold of $5cm, 5^\circ$. For Kings College, we report the median translation (cm), rotation ($^\circ$) and pose accuracy at the error threshold of $25cm, 2^\circ$. We also provide the time of rendering a patch.}
	\label{tab:ablation}
\end{table}

%% file: sections/conclusion.tex
\section{Conclusion}
\label{sec:conclusion}

In this paper, we propose a new method of applying NeRFs to visual localization task. In detail, we introduce the explicit geometric map (EGM) and implicit learned map (ILM) to provide sparse keypoints and rendered patches to build sparse matches between query and rendered images. By adopting sparse rendering from sparse points provided by EGM, our approach avoids time-consuming full image rendering. With ILM represented by NeRFs, our method discards the memory-consuming 2D descriptors. Therefore, our system is more efficient. However, the accuracy on large-scale Aachen dataset is still limited compared to state-of-the-art methods. We hope this work could be a baseline and the more researchers can make it better in the future.